\documentclass[lettersize,journal]{IEEEtran}
\usepackage{amsmath,amsfonts}
\usepackage{algorithmic}
\usepackage{algorithm}
\usepackage{array}
\usepackage[caption=false,font=normalsize,labelfont=sf,textfont=sf]{subfig}
\usepackage{textcomp}
\usepackage{stfloats}
\usepackage{url}
\usepackage{verbatim}
\usepackage{graphicx}
\usepackage{cite}
\usepackage{booktabs}
\usepackage{multirow}
\usepackage{orcidlink} 
\hypersetup{hidelinks} 

\begin{document}
\title{Reliable Multi-Modal Object Re-Identification via Modality-Aware Graph Reasoning}

\author{Xixi Wan$^{\orcidlink{0009-0006-6254-6713}}$, Aihua Zheng\(^{*\orcidlink{0000-0002-9820-4743}}\),  Zi Wang$^{\orcidlink{0000-0002-8001-0318}}$, Bo Jiang\(^{*\orcidlink{0000-0002-6238-1596}}\), Jin Tang$^{\orcidlink{0000-0001-8375-3590}}$, and Jixin Ma$^{\orcidlink{0000-0001-7458-7412}}$

\thanks{
This research is supported in part by the National Natural Science Foundation of China (62372003), the University Synergy Innovation Program of Anhui Province (GXXT-2022-036), the Natural Science Foundation of Anhui Province (2308085Y40, 2408085J037), Key Technologies R \& D Program of Anhui Province (202423k09020039), and the Key Laboratory of Intelligent Computing \& Signal Processing, Ministry of Education, Anhui University (2024A004). (\(^*\)Corresponding authors are Aihua Zheng and Bo Jiang.)
}

\thanks{
A. Zheng and X. Wan are with the Information Materials and Intelligent Sensing Laboratory of Anhui Province, Anhui Provincial Key Laboratory of Security Artificial Intelligence, School of Artificial Intelligence, Anhui University, Hefei, 230601, China
(e-mail: ahzheng214@foxmail.com; xixiwan11@163.com).
}
\thanks{
Z. Wang is with the School of Biomedical Engineering, Anhui Medical University, and the Key Laboratory of Intelligent Computing \& Signal Processing, Ministry of Education, Anhui University, Hefei, China. (e-mail: ziwang1121@foxmail.com).
}
\thanks{
B. Jiang and J. Tang are with Anhui Provincial Key Laboratory of Multimodal Cognitive Computation, School of Computer Science and Technology, Anhui University, Hefei, 230601, China. 
(e-mail: zeyiabc@163.com; ahu\_tj@163.com).
}
\thanks{
Jixin Ma is with the School of Computing and Mathematical Sciences, University of Greenwich, London SE10 9LS, UK. 
(e-mail: j.ma@greenwich.ac.uk).
}

}

\markboth{Journal of \LaTeX\ Class Files,~Vol.~14, No.~8, August~2021}%
{Shell \MakeLowercase{\textit{et al.}}: A Sample Article Using IEEEtran.cls for IEEE Journals}

\maketitle

\begin{abstract}
Multi-modal data provides abundant and diverse object information, crucial for effective modal interactions in Re-Identification (ReID) tasks. However, existing approaches often overlook the quality variations in local features and fail to fully leverage the complementary information across modalities, particularly in the case of low-quality features. 
In this paper, we propose to address this issue by leveraging a novel graph 
reasoning model, termed the Modality-aware Graph Reasoning Network (MGRNet). 
Specifically, we first construct modality-aware graphs to enhance the extraction of fine-grained local details by effectively capturing and modeling the relationships between patches.
Subsequently, the selective graph nodes swap operation is employed to alleviate the adverse effects of low-quality local features by considering both local and global information, enhancing the representation of discriminative information. 
Finally, the swapped modality-aware graphs are fed into the local-aware graph reasoning module, which propagates multi-modal information to yield a reliable feature representation. 
Another advantage of the proposed graph reasoning approach is its ability to reconstruct missing modal information by exploiting inherent structural relationships, thereby minimizing disparities between different modalities.
Experimental results on four benchmarks (RGBNT201, Market1501-MM, RGBNT100, MSVR310) indicate that the proposed method achieves state-of-the-art performance in multi-modal object ReID. 
The code for our method will be available upon acceptance.

\end{abstract}

\section{Introduction}
\begin{figure}
\centering
\captionsetup{font=small}
\includegraphics[width=1.0\linewidth]{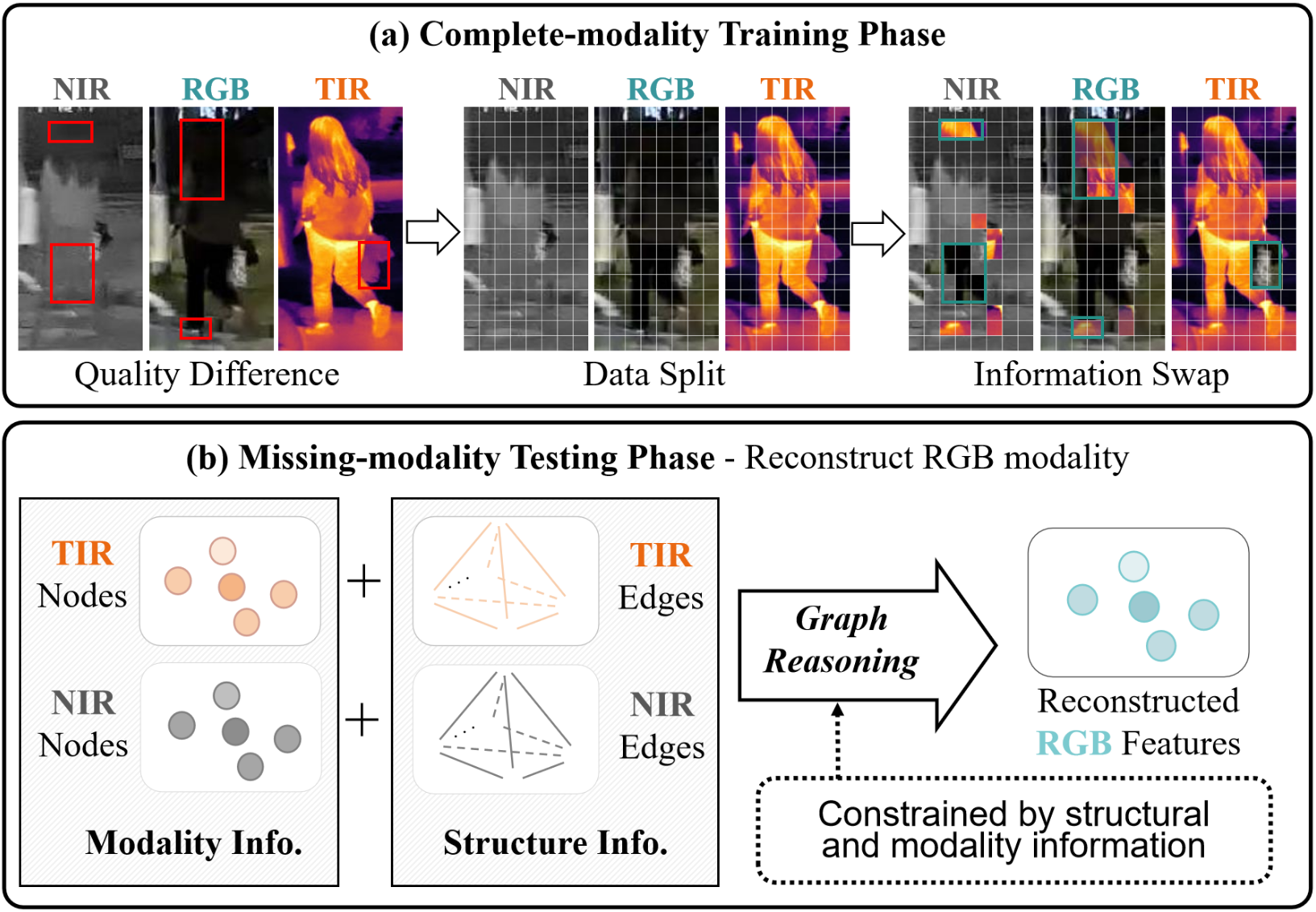}
\caption{(a) Due to quality differences in local features among modalities, we first split the data to obtain more detailed local information, and then perform an information swap (achieved by GRMI described in Sec~\ref{sec:GRMI}). (b) When TIR is missing in the testing phase, we leverage graph reasoning trained with constraints from modality and structural information to restore features, combining existing RGB and NIR features (nodes) and their relationships (edges), achieved by GRMM described in Sec~\ref{sec:GRMM}.
}
\label{motivation}
\end{figure}
\IEEEPARstart{R}{ecently}, multi-modal data has gained prominence as a promising problem in computer vision, particularly for the task of object ReID~\cite{9336268,10400493,Wang2024MambaPro}.
Compared with single-modal and cross-modal data~\cite{10081455,Dai2018CrossModalityPR,9011001,10.1145/3343031.3351006,He_2021_ICCV,10807364}, multi-modal data can capture object features more comprehensively by integrating information from different data sources, which is beneficial for practical object ReID scenario. For example, Near Infrared (NIR) images can provide more visible information
especially in low illumination, while Thermal Infrared (TIR) images have a strong ability to penetrate haze and smog~\cite{Li2020InfraredVisibleCP,Zheng2021RobustMP,wang2024demo,wang2025IDEA,10364832}.
Previous works have demonstrated the effectiveness of multiple modalities in enhancing the performance of the object ReID task~\cite{Li2020MultiSpectralVR,Ye2020DynamicDA,Zheng2021RobustMP,Wang2022InteractEA,Crawford2023UniCatCA,Zheng2023DynamicEN,ZHENG2023101901,Wang2023TOPReIDMO,Wang2024HeterogeneousTT,10654953,wang2025IDEA}. 
However, we observe that existing multi-modal methods often rely on global feature extraction to exploit the complementary information between different modalities~\cite{Li2020MultiSpectralVR,Zheng2023DynamicEN,ZHENG2023101901,Crawford2023UniCatCA}, which ignores mining fine-grained local cues for the multi-modal ReID task. To address this issue, part-based methods are introduced to the multi-modal object ReID by simply dividing features either randomly or based on object part positions inferred from vision encoders ~\cite{10654953,10772090,Wang2024MambaPro,wang2025IDEA}.
\emph{Despite these part-based methods can extract and utilize local information, they fail to consider the quality differences between local features of different modalities, which are ineffective for low-quality local features, as shown in Fig.~\ref{motivation} (a).}
Additionally,
in real-world scenarios, the issue of missing modalities is common and often unavoidable. 
For example,
Zhu \emph{et al.}~\cite{zhu2017gan} and Wang \emph{et al.}~\cite{Wang2023TOPReIDMO, wang2024demo}  propose image and feature reconstruction to compensate for the absence of images, respectively.
Wang \emph{et al.}~\cite{9156763} utilize zero padding to reconstruct information. 
\emph{However, existing works mainly focus on feature- or image-level reconstruction while neglecting the essential structural relationships and dependencies among features of the missing modality. 
This leads to the generation of missing modalities with inconsistent structure and semantics, impacting the model's performance.}

To address the above limitations,  we propose a new Modality-aware Graph Reasoning
Network (MGRNet) that can effectively boost information interactions and recover missing modalities for multi-modal object ReID. 
Specifically, 
the modality-aware graphs are first utilized to facilitate the extraction of important local details by modeling the relationships between patches for multi-modal data. 
Obviously, existing low-quality information somewhat increases the difficulty of learning spectral image features. Meanwhile, current approaches often struggle to fully obtain the complementarity and dependencies between modalities in Fig.~\ref{motivation} (a). These problems prompt us to propose a new multi-modal fusion method for object ReID via selective graph nodes swap operation.
This operation is designed to effectively mitigate the impact of low-quality local features by considering both local and global information, thereby enhancing the discriminative information. 
As shown in Fig.~\ref{motivation} (a), swapping local features addresses issues related to low-quality features, thus improving the effectiveness of modal representations.
Subsequently, we feed the swapped modality-aware graphs into the local-aware graph reasoning module to achieve multi-modal information propagation, thus yielding reliable feature representations. 
Another key aspect of the proposed graph reasoning is that it can recover the features of missing modalities by exploiting their structural relationships, thereby minimizing the undesired disparity between different modalities. 
This recovered loss is intended to reduce the gap between the reconstructed and real representations, as well as to minimize the disparity between modalities. 
During the testing phase, the reconstruction module can be directly utilized to generate features for the missing modality, as shown in Fig.~\ref{motivation} (b). 

Overall, MGRNet can capture rich, fine-grained local features while reducing the influence of low-quality local tokens by considering both local and global information, jointly promoting information interaction of object. 
To the best of our knowledge, this is the first attempt to leverage graph reasoning for both modal interaction and missing problems in multi-modal object ReID, which achieves outstanding performance on common object datasets. 
The main contributions of this work are summarized as,
\begin{itemize}
\item[$\bullet$] \noindent We propose the Modality-aware Graph Reasoning Network (MGRNet) to jointly capture the crucial structure relationships and complementary information among different modal tokens, solving modal interaction and missing problems for multi-modality object ReID.
\item[$\bullet$] \noindent We introduce multiple modality-aware graphs to incorporate structural information of local features and design the selective graph nodes swap operation to effectively alleviate the impact of low-quality local features. 

\item[$\bullet$] \noindent
The MGRNet inherently possesses the ability to reconstruct the features of missing modalities based on their structural relationships, as well as to minimize the disparity between modalities. 

\item[$\bullet$] \noindent 
Extensive experiments on four public benchmarks show that MGRNet achieves superior performance over state-of-the-art approaches for multi-modal object ReID, validating its reliability and effectiveness.

\end{itemize}
\section{Related Works}
\subsection{Multi-Modal Object ReID}
At present, existing research has made significant progress for multi-modal object ReID. For example, in works~\cite{Zheng2021RobustMP,Wang2023TOPReIDMO,10654953,10005795}, they utilize and fuse multiple source data to provide a more comprehensive interaction of modalities. Among that,
some methods leveraging global-based methods have been proposed \cite{Zheng2023DynamicEN,ZHENG2023101901,Wang2024HeterogeneousTT}.
CCNet \cite{ZHENG2023101901} proposes CDC loss to solve discrepancies in both modalities and samples, generating discriminative multi-modal feature representations for vehicle ReID. 
HTT \cite{Wang2024HeterogeneousTT} proposes to utilize CIM loss to enlarge the differentiation between distinct identity samples and then apply MTT to optimize the generalization capabilities of the model. 
DeMo \cite{wang2024demo} first decouples multi-modal features to obtain modal-specific and -shared information. Then, an Attention-Triggered Mixture of Experts (ATMoE) is leveraged to enhance modal interactions via attention-guided operation.
However, the above methods only focus on the complementarity and dependencies of global information while overlooking the deep interaction of local information. 
For multi-modal object ReID, it is well-known that local features provide rich, fine-grained information about objects information, which is essential for effective feature interaction and reconstruction. 
Thus, part-based multi-modal methods
are proposed~\cite{Zheng2021RobustMP, Wang2022InteractEA, 10654953, Wang2023TOPReIDMO}. PFNet\cite{Zheng2021RobustMP} proposes a progressive fusion network to effectively fuse different spectral features, including RGB, NIR, and TIR for object ReID task. 
IEEE~\cite{Wang2022InteractEA} proposes a relation-based embedding module to embed the global information into fine-grained local features, boosting feature representations for multi-modality ReID.
EDITOR~\cite{10654953}  introduces object-centric tokens for multi-modal ReID. This method proposes SFTS and HMA to select and aggregate multi-modal tokenized features. 
TOP-ReID~\cite{Wang2023TOPReIDMO} proposes a Token Permutation Module (TPM). This module can align multi-spectral images and utilize the current global token to perceive the local token of other modalities. 
IDEA~\cite{wang2025IDEA} adaptively generates sampling positions by aggregated multi-modal information to facilitate the interaction between global features and local information.

Although these methods can extract and use local information. They either adopt random segmentation or divide features into multiple parts based on object parts, ignoring the quality difference between local features of different modalities, thereby creating certain noise for the extracted features.

\subsection{Graph-Based Object ReID}
Graph-Based methods have been widely applied in object ReID which can model relationships between single spectrum images and promote feature representations ~\cite{10.1145/3394171.3413578,9859883,9522697,Liu2021PrGCNPP,10684797}. For instance, 
Nguyen \emph{et al.}~\cite{9522697} propose a graph-based person signature, fusing detailed person descriptions and visual features into a graph.
Jiang \emph{et al.}~\cite{9479704} propose PH-GCN to tackle the single-modal person ReID problem, offering a unified solution that effectively integrates local, global, and structural feature representations.
He \emph{et al.}~\cite{He2021PGGANetPG} propose the PGGANet method, utilizing a self-adaptive graph attention convolution to learn the contribution matrix of local information.
Sun \emph{et al.}~\cite{10400493}  propose a polymorphic masked wavelet graph convolutional network to disentangle content and degradation features of cross-modality images.
Lv \emph{et al.}~\cite{10494037} propose a novel edge weight-embedding graph convolutional network that embeds human joints and bones into the feature representation of object ReID.
While these methods achieve relatively good results in single-spectrum object ReID, they mainly emphasize interactions for global and local features,  neglecting the quality difference of local features and the complementary information cross modalities.

Notably, unlike the above methods, our MGRNet is the first to propose utilizing graph reasoning to exploit both modal interaction and missing problems by considering both global and local features in multi-modal object ReID. We first design a Graph Reasoning on Modal Interaction (GRMI) strategy to adaptively learn information between patches of multi-modal images, meanwhile presenting a novel Graph Reasoning on Missing Modality (GRMM) strategy to compensate for missing modal information and reduce the difference between modalities. 
\section{The Proposed Method}
\subsection{Overview}
In this section, we present a novel method to enhance information interaction and reconstruct missing modality called Modality-aware Graph Reasoning Network (MGRNet) for multi-modal object ReID. This method consists of four main parts: (1) Initial Feature Extraction,  (2) Graph Reasoning on Modal Interaction, (3) Global-aware Multi-Head Attention, and (4) Graph Reasoning on Missing Modality, as illustrated in Fig.~\ref{overall}.
\begin{itemize}
\item[$\bullet$] \noindent{Initial Feature Extraction.} To capture the distinctive information of each modality, we leverage a multi-branch vision encoder to extract rich modal features.
\item[$\bullet$] \noindent{Graph Reasoning on Modal Interaction (GRMI) strategy.} 
To extract the discriminative information, we design a novel graph reasoning on modal interaction strategy to capture local details while alleviating the impact of low-quality local features by considering both local and global features.
\item[$\bullet$] \noindent{Global-aware Multi-Head Attention.} To obtain better global information, we introduce the global-aware multi-head attention module to aggregate rich local information from different patches into global features.
\item[$\bullet$] \noindent{Graph Reasoning on Missing Modality (GRMM) strategy.}
To address modal missing, we propose the graph reasoning on missing modality strategy to recover missing modal information while reducing the disparity between multi-modality data.
\end{itemize}
Finally, a comprehensive train loss is utilized to optimize the entire proposed method. Below, we will introduce the main parts in detail.

\subsection{Initial Feature Extraction}
We adopt multi-branch backbones for extracting feature representations from multi-modal data, given their excellent performance for multi-modal images in object ReID~\cite{Wang2024HeterogeneousTT,10654953,wang2025IDEA}. 
To preserve special information about each modality and extract better initial features of different modalities, the multi-branch backbones are non-shared. Formationally, this process can be defined as follows,
\begin{equation}\label{ViT1}
    X^{m} = \{X^{m}_g, X^{m}_l\}  = {{\xi}^{m}}(I^{m}),
\end{equation}
where ${\xi}^{m}$ denotes the vision encoder (ViT or CLIP)~\cite{Wang2024HeterogeneousTT,Wang2023TOPReIDMO,wang2025IDEA}.  $I^{m} \in \mathbb{R}^{256 \times 128 \times 3}$ is the input images for the $m$-th modality,  with $m \in \{N, R, T\}$ where $N$, $R$, and $T$ refer to the NIR, RGB, and TIR modalities, respectively. $X^{m}$ $\in \mathbb{R}^{(P+1) \times D}$ is the extracted features  by ViT for the $m$-th modality.
$ X^{m}_l$ 
$\in \mathbb{R}^{P \times D} $ and $ X^{m}_g$ 
$\in \mathbb{R}^D $ represent patch tokens 
and class tokens of input images, respectively. $P = 128$ is number of local tokens and $D$ denotes the
obtained dimensions of embedding tokens.

\subsection {Graph Reasoning on Modal Interaction}\label{sec:GRMI}
\textbf{Modality-aware Graph Learning (MGL).} 
To fully encode the local features obtained above of multi-modal information, we build a modality-aware graph $G(V^{m}, E^{m})$ for each modality. For the $m$-th modality, 
the nodes $V^{m}$ of our modality-aware graph are represented as the set of local features of images. 
To be specific, let 
${{X}_{l}^{m}} = \{X^{m}_{l_1}, X^{m}_{l_2} \cdots X^{m}_{l_P}\}
 \in \mathbb{R}^{P \times D} $ denotes the collection features of all patch where $X^{m}_{l_p}$ means the feature representation of the $p$-th patch node in the $m$-th modality. 
The edges $E^{m}$ of our modality-aware graph connect different patches, utilizing an adjacency matrix $A^{m}$ to represent 
structural relationships among local features.  
It is learned dynamically  by  computing the Euclidean Distance~\cite{10.1145/3459637.3482261} to construct relationships as,
\begin{equation}\label{dist0}
    {{D}}_{l_{ij}}^{m} = {\psi}(X^m_{l_i}, X^m_{l_j}),
\end{equation}
where $X^m_{l_i}$ and $X^m_{l_j}$ represent the feature vectors of the $i$-th and $j$-th patches  for the $m$ modality, respectively. And $\psi$ is the Euclidean Distance~\cite{10.1145/3459637.3482261}.
Next, to
learn more effective graphs, we also introduce two learnable parameters $\alpha$ and $\beta$ into  ${{A}}_{l_{ij}}^{m}$, which is defined as follows,
\begin{equation}\label{dist1}
    {{A^{m}_{l_{ij}}}} = 1 - \sigma\big(({{D}}_{l_{ij}}^{m} + \alpha) \times \beta \big),
\end{equation}
where $\sigma$ is the sigmoid activation function. 

\begin{figure}[t]
\centering
\includegraphics[width=0.95\columnwidth]{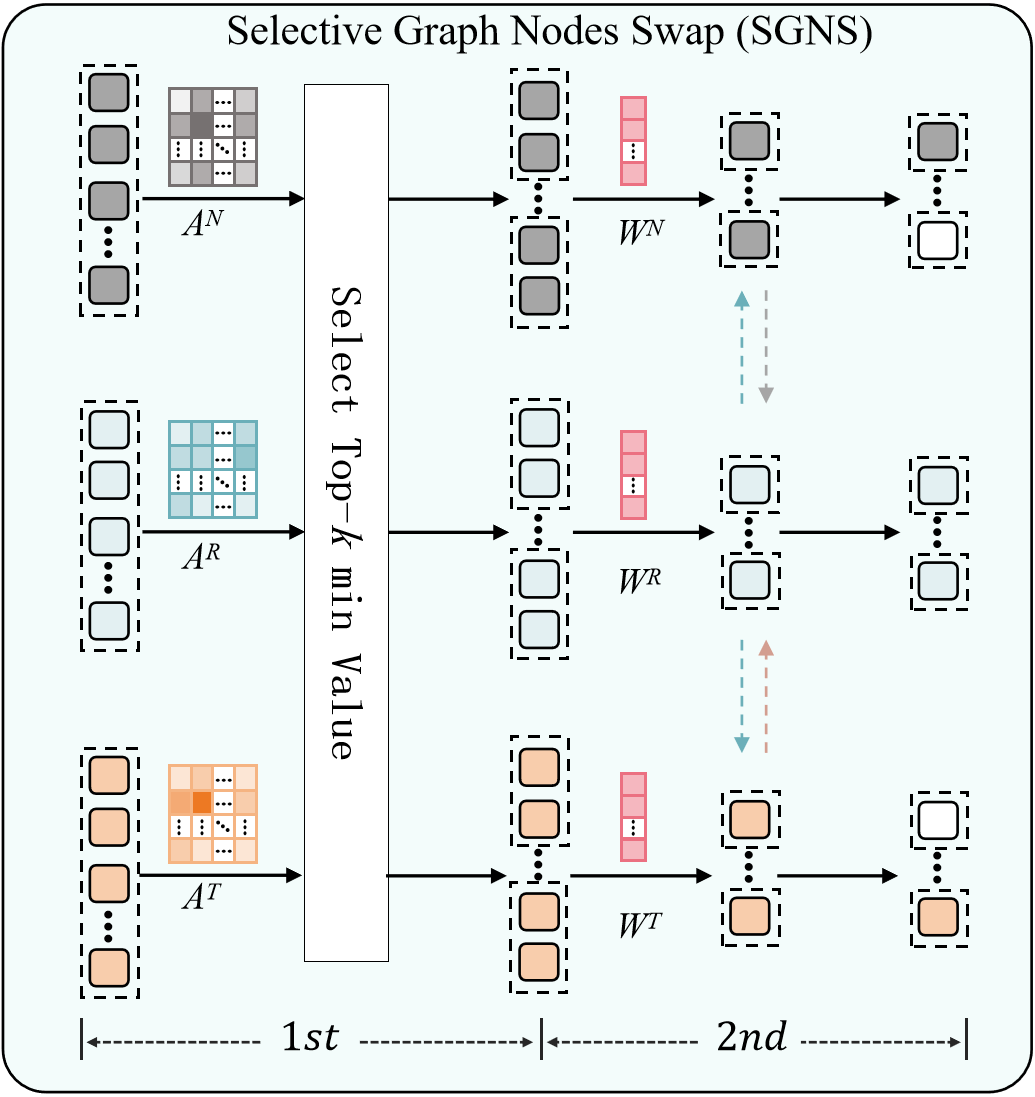} 
\caption{The process of multiplying selective graph nodes swap by considering both local and global information.}
\label{part}
\end{figure}

\textbf{Selective Graph Nodes Swap (SGNS).} 
To further promote interaction among obtained local features, the Selective Graph Nodes Swap (SGNS) operation is designed to reduce the impact of low-quality local details between modalities through considering both local and global information as shown in Fig.~\ref{part}. 
This operation consists of two main steps.
\textbf{Firstly}, the Top-$k$ method is employed to select $k$ small values for edges of each graph and then find the poor patches corresponding. \textbf{Secondly}, based on the above method, we also introduce global tokens to select more precise local nodes about low-quality patches. Specifically, The process first calculates the similarity between the global feature and each local feature as follows, 
\begin{equation}\label{dist2}
    {{W}}^{m} = 1- \phi\big({\psi}(X^m_{g}, X^m_{l})\big),
\end{equation}
where 
${W}^{m} =\{{w}^{m}_1, {w}^{m}_2 \cdots {w}^{m}_P \}\in \mathbb{R}^P$
denotes the strength of the relationship between global and local features. $\phi$ is the softmax of the non-linear activation function.
And then ${W}^{m}$ is applied to screen the first step to get the poor nodes, which is beneficial for obtaining lower-quality patches. 
\begin{figure*}
\centering
\includegraphics[width=2\columnwidth]{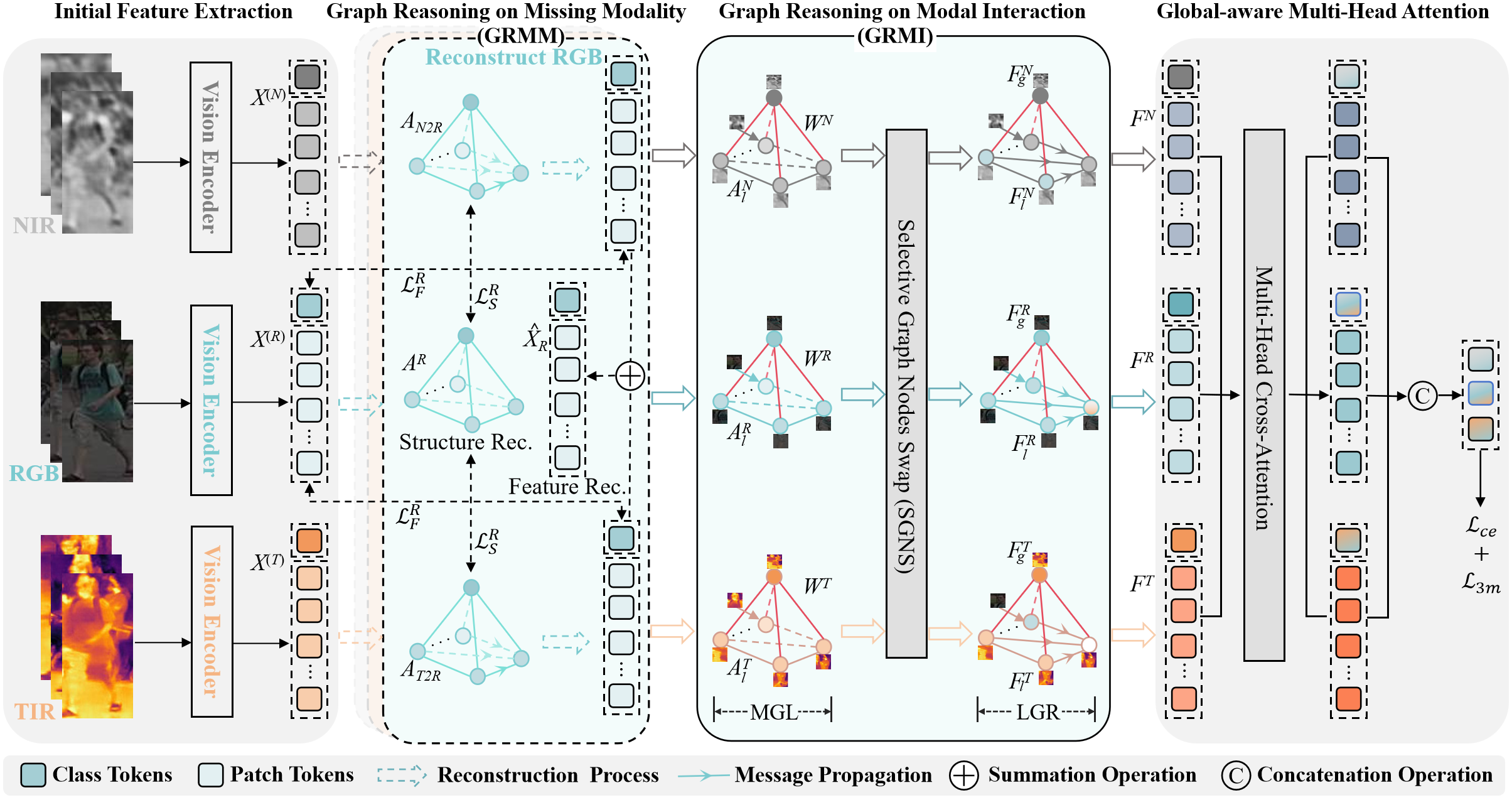}
\caption{The overall network structure of the proposed MRGNet.
For complete multi-modal training and testing, initial feature extraction first employs the multi-branch vision encoders on the multi-modal images to obtain the initial features. Secondly, graph reasoning on modal interaction is employed to alleviate low-quality tokens of each modality. Finally, the enhanced features are generated with global-aware multi-head attention and the fused features are fed into the classifiers to get the ReID results. Furthermore, graph reasoning on missing modality strategy is designed to restore features based on their structural relationships for the missing modality problems.
}
\label{overall}
\end{figure*}
As we know, local features perform differently across modalities, resulting in varying quality. Thus, local patches between RGB, NIR, and TIR will be performed through exchange based on their respective quality; for example, the poor nodes of RGB will be replaced by mean features of NIR and TIR. Finally, we obtain the updated nodes feature of local features. 
Note that we replace the inferior patches by leveraging better patches of other modalities while taking into account the correlation between patches. 
So the updated $X^{m}_{l_i}$ is defined as,
\begin{equation}\label{swap3}
    X^{R}_{l_i} =Swap \big( X^{R}_{l_i}, \frac{1}{2}(X^{N}_{l_i} + X^{T}_{l_i})\big),
\end{equation}
\begin{equation}\label{swap1}
    X^{N}_{l_i} =Swap \big( X^{N}_{l_i}, X^{R}_{l_i}\big),
\end{equation}
\begin{equation}\label{swap2}
    X^{T}_{l_i} =Swap \big( X^{T}_{l_i}, X^{R}_{l_i}\big).
\end{equation}
In addition, to avoid the exchange of lower-quality patches, we also designed to determine whether the swapped patch is a poor patch.  If so, we re-initialize the current patch token that is set all-zero matrix and then learn the feature expression of the current patch through the local neighbors of the current patch, otherwise, we exchange this patch as Fig.~\ref{part}. This design can not only facilitate the learning of information from other modalities but also mitigate the impact of low-quality local features by considering both local and global information.

\textbf{Local-aware Graph Reasoning (LGR).}
 Based on the above SGNS operation, we obtain enhanced node representations of each modality-aware graph. Accordingly, the multi-layer local-aware graph reasoning module is employed to learn local patches representation of multi-modal data for better quality node patches. The message propagation rule is defined as follows,
\begin{equation}\label{GCN}
    {{{{F}}_l^{(m,\tilde{l}+1)}}={\delta({{{{A}}^{m}_l}{{{F}_l^{{(m,\tilde{l})}}}}{{{\Theta}}^{(m,\tilde{l})}})}}},
\end{equation}
where $\tilde{l} = 0, 1 \cdots \tilde{L}-1$ denotes the $\tilde{l}$ layer of GCN and ${F}_l^{(m,0)} = {X}_l^{m}$. ${{{\Theta}}^{(m,\tilde{l})}}$ is the learnable transformation matrix and  ${{F}}_l^{(m,\tilde{L})}$  is briefly represented as ${{F}}_l^{m}$.  $\delta$ denotes the non-linear activation function ReLU.

\subsection{Global-aware Multi-Head Attention}
Let ${{{F}}_l^{m}} = \{
{{{F}}_{l_1}^{m}}, {{{F}}_{l_2}^{m}} 
\cdots {{{F}}_{l_P}^{m}}\}$ 
denotes the obtained patches via the above graph reasoning on modal interaction strategy.  We introduce the global-aware multi-head attention module which can capture richer node representations by increasing the interaction of global and local information. Formationally, We apply  a linear project layer  to obtain a query matrix 
$Q^{m} \in \mathbb{R}^{D}$
for global token ${{{F}}_g^{m}} = {{{X}}_g^{m}}$ and different linear project layer to get key matrix 
$K^{m} \in \mathbb{R}^{P \times D}$
and value matrix 
$V^{m} \in \mathbb{R}^{P \times D}$
for local token ${{{F}}_l^{m}}$. Then, 
multi-head attention with $H$ heads is utilized to aggregate local information from different patches into global feature representations. This interaction operation is defined in the $h$-th head as follows,
\begin{equation}\label{att}
    {{\hat{F}}^{(m,h)}}= \phi[\frac{{{{Q}}^{(m,h)}}{{{({K}}^{(m,h)})}^T}}{\tau}]{{{V}}^{(m,h)}},
\end{equation}
where $\tau $ is the scale factor and $\phi$ is the softmax of the non-linear activation function. ${\mathbf{\hat{F}}^{(m,h)}}$ represents obtained output feature of the 
the $h$-th head for the $m$-th modality. To aggregate the information from all heads, we utilize the concatenation operation to get a new
class token as,
\begin{equation}\label{concat}
    {\mathbf{\hat{F}}^{m}} = \mathrm{Con}\big({\hat{F}^{(m,1)}} \cdots {\hat{F}^{(m,H)}}\big).
\end{equation}
Finally, we concatenate features  of all modalities to obtain the fused features $\mathbf{{Z}}$ as follows,
\begin{equation}\label{concat}
    \mathbf{Z} = \mathrm{Con}(\mathbf{\hat{F}}^{N}, \mathbf{\hat{F}}^{R}, \mathbf{\hat{F}}^{T}).
\end{equation}

\subsection{Graph Reasoning on  Missing Modality}
\label{sec:GRMM}
To address the modality-missing issue in the real world, the MGRNet also design the Graph Reasoning on  Missing Modality (GRMM) strategy to compensate for missing modal information and reduce the difference between modalities as Fig.~\ref{overall}. 
GRMM first applies feature reconstruction to enhance feature representations, then takes advantage of structure reconstruction to learn the relationships of corresponding tokens.
To be more specific, assuming that the RGB modality is missing, we leverage the existing NIR and TIR modalities to recover the tokens of the missing RGB modality. We first compute the original structure relationships  between all real tokens of each modality as follows,
\begin{equation}\label{dist3}
    {{A}}_{ij}^{m} = 1 - \sigma({{D}}_{ij}^{m}), {{D}}_{ij}^{m} =  \psi(X^{m}_i, X^{m}_j),
\end{equation}
where $X^{m}_{i}$ and $X^{m}_{j}$ represent the feature vectors of the tokens $i$ and $j$ respectively for the $m$-th modality. $\psi$ is the Euclidean Distance~\cite{10.1145/3459637.3482261}. $\sigma$ is the sigmoid activation function.
Next, we construct dynamically reconstructed structure relationships ${A}_{N2R}$, ${A}_{T2R}\in \mathbb{R}^{(P+1) \times (P+1)}$ as follows,
\begin{equation}\label{dist4}
    {{A}}_{N2R} = 1 - \sigma(t_N{{D}}^{N}),
\end{equation}
\begin{equation}\label{dist5}
    {{A}}_{T2R} = 1 - \sigma(t_T{{D}}^{T}),
\end{equation}
where $t_N$ and $t_T$  are two learnable hyperparameters. And then equipped with a layer-wise GCN to propagate messages for recovering feature $\hat{X}_{N2R}$ and $\hat{X}_{T2R}$. This recovered process is defined as,
\begin{equation}\label{GCN1}
    {{{\hat{X}}_{N2R}^{\hat{l}+1}}={\delta({{{{A}}_{N2R}}{{{\hat{X}}_{N2R}^{\hat{l}}}}{{{\Theta}}^{(N,\hat{l})}})}}},
\end{equation}
\begin{equation}\label{GCN2}
    {{{\hat{X}}_{T2R}^{\hat{l}+1}}={\delta({{{{A}}_{T2R}}{{{\hat{X}}_{T2R}^{\hat{l}}}}{{{\Theta}}^{(T,\hat{l})}})}}},
\end{equation}
where 
$\hat{l}=0, 1 \cdots \hat{L}-1$ denotes the $\hat{l}$ layer of GCN  and $\hat{X}^{0}_{N2R} = X^{N}, \hat{X}^{0}_{T2R} = X^{T} \in \mathbb{R}^{(P+1) \times D}$. 
$\Theta^{(N,\hat{l})}$ and $\Theta^{(T,\hat{l})}$ are
the layer-wise trainable transformation matrices. The outputs via  $\hat{L}$ layer-wise GCN are $\hat{X}^{\hat{L}}_{N2R}, \hat{X}^{\hat{L}}_{T2R}$ which are briefly denoted as $\hat{X}_{N2R}, \hat{X}_{T2R}$. 
In the absence of the RGB modality, we dynamically generate its token features using the information from NIR and TIR modalities as follows,
\begin{equation}\label{GCN}
    {{{\hat{X}^R}}= {\frac{1}{2}({\hat{X}_{N2R}}} + {{\hat{X}_{T2R}}})},
\end{equation}
Then, the reconstructed token features ${\hat{X}^R}$, ${X}^{N}$, $X^T$ are fed into graph reasoning on modal interaction strategy to jointly fusion of multi-modal data for ReID.
The proposed GRMM strategy not only solves the limitation of the receptive field of CNN architecture by convolution on the graph structure but also effectively solves the challenge of missing modality in the ReID task.

\subsection{Train Loss}
To promote the reconstruction ability, the recovered features are constrained via leveraging both features reconstruction loss ${\mathcal{L}}^R_F$ and structure reconstruction loss $\mathcal{L}^R_S$ as follows,
\begin{equation}\label{loss1}
    \small
    {\mathcal{L}}^R_F = \frac{1}{P+1} {\sum_{p=1}^{P+1}} \|{\hat{X}_{N2R}}-{X^{R}}\|^2 + \|{\hat{X}_{T2R}}-{X^{R}}\|^2,
\end{equation}

\begin{equation}\label{loss2}
    \small
    {\mathcal{L}}^R_S = \frac{1}{P+1} {\sum_{p=1}^{P+1}} \|{A_{N2R}}-{A^{R}}\|^2 + \|{A_{T2R}}-{A^{R}}\|^2,
\end{equation}
\begin{equation}\label{loss3}
    {\mathcal{L}}^{R} = {\mathcal{L}}^R_F + {\mathcal{L}}^R_S.
\end{equation}
We observe that the constraints can also reduce the gap between the generated RGB modality and other modalities. Similarly, if other modalities are missing, reconstruction can also be achieved to obtain a smaller modal distribution gap. Thus, the proposed GRMM strategy is training under the constraints of multi-modality reconstruction loss $\mathcal{L}_{MR}$  as,
\begin{equation}\label{loss4}
    {\mathcal{L}}_{MR} = {\mathcal{L}}^{R} + {\mathcal{L}}^{N} + {\mathcal{L}}^{T}.
\end{equation}

In the training phase, our train loss contains several parts, including Vision Encoders, Graph Reasoning on Missing Modality,  Graph Reasoning on Modal Interaction, and Global-aware Multi-Head Attention. We optimize the entire network by utilizing the above multi-modality reconstruction loss, multi-modality margin~\cite{Wang2022InteractEA} and cross-entropy loss~\cite{7780677}, minimizing the sum of all losses as follows,
\begin{equation}\label{loss5}
    {\mathcal{L}} = {\mathcal{L}}_{MR}  + {\mathcal{L}}_{ce} + {\mathcal{L}}_{3m},
\end{equation}
${\mathcal{L}}_{3m}$ is removed for the proposed MGRNet of CLIP-based vision encoders.
This overall network is optimized in an end-to-end manner.

\section{EXPERIMENTAL RESULTS AND ANALYSIS}
In this section, we evaluate the effectiveness of the proposed MGRNet on four commonly used datasets and compare it with some other related works.

\begin{table*}[!ht]
\caption{Comparison with state-of-the-art methods on the person ReID datasets(in \%). The symbol $^\ast$ indicates ViT-based methods, while $^\dagger$ represents Clip-based methods for our proposed MGRNet.} 
\label{mytable1}
\centering
\resizebox{2.05\columnwidth}{!}{
\begin{tabular}{c|c|c|c|cccc|cccc}
\toprule
&{\multirow{2}{*}{\textbf{Methods}}}
&\multirow{2}{*}{\textbf{Publication}}&\multirow{2}{*}{\textbf{Structure}}
&\multicolumn{4}{c|}{\textbf{RGBNT201}}
&\multicolumn{4}{c}{\textbf{Market1501-MM}}\\
\cmidrule{5-8}
\cmidrule{9-12}
&&&&\textbf{mAP} & \textbf{R-1}&\textbf{R-5} & \textbf{R-10}&\textbf{mAP} & \textbf{R-1}&\textbf{R-5} & \textbf{R-10}\\
\midrule
\multirow{4}{*}{\rotatebox{0}{\textbf{Single}}}
&{HACNN}~\cite{Li_2018_CVPR}&{CVPR18}&CNN&21.3&19.0& 34.1&42.8&42.9&69.1&86.6&92.2 \\
&{MLFN}~\cite{8578323}& {CVPR18}&CNN&26.1&24.2&35.9&44.1&42.7&68.1&87.1&92.0 \\
&{OSNet}~\cite{9011001}& {ICCV19}&CNN&25.4&22.3& 35.1&44.7&39.7&69.3&86.7&91.3 \\
&{TransReID}~\cite{He_2021_ICCV}& {ICCV21}&ViT&63.8&65.8&78.5&83.9&73.0&88.9&95.8&97.6 \\
\midrule
\multirow{12}{*}{\rotatebox{0}{\textbf{Multi}}}
&{HAMNet}~\cite{Li2020MultiSpectralVR}& {AAAI20}&CNN&27.7&26.3& 41.5&51.7&60.0& 82.8& 92.5 &95.0 \\
&{PFNet}~\cite{Zheng2021RobustMP}& {AAAI21}&CNN&38.5& 38.9& 52.0 &58.4&60.9 &83.6& 92.8& 95.5 \\
&{IEEE}~\cite{Wang2022InteractEA}& {AAAI22}&CNN&
 46.4& 47.1& 58.5& 64.2&64.3& 83.9 &93.0 &95.7 \\
&{UniCat}~\cite{Crawford2023UniCatCA}& {NIPSW23}&ViT&
 57.0& 55.7& -& -&-& - &- &- \\
&{EDITOR}~\cite{10654953}&{CVPR24}&ViT&65.7&68.8&82.5&89.1&77.4&90.8&96.8&98.3 \\
&{RSCNet}~\cite{10772090}& {TCSVT24}&ViT&68.2 &72.5&-&-&-&-&-&-\\
&{HTT}~\cite{Wang2024HeterogeneousTT}& {AAAI24}&ViT&71.1&73.4&83.1&87.3&67.2& 81.5& 95.8& 97.8 \\
&{TOPReID}~\cite{Wang2023TOPReIDMO}& {AAAI24}&ViT&72.3&76.6&84.7&89.4&\underline{82.0}	&\underline{92.4}&	\underline{97.6}&\underline{98.6}\\
&{DeMo}~\cite{wang2024demo}& {AAAI25}&ViT&\underline{73.7}&\underline{80.5} &\underline{88.3} &\underline{91.5}&78.0&90.7&96.8&98.2\\
&{\textbf{MGRNet$^\ast$}}& \textbf{Ours}&ViT&\textbf
{78.4}&\textbf{82.5}&\textbf{90.9}&\textbf{95.2}&\textbf{84.4}&\textbf{94.0}&\textbf{98.2}&\textbf{98.9}\\
\cmidrule{2-12}
&{MambaPro}~\cite{Wang2024MambaPro}& {AAAI25}&CLIP&78.9&\underline{83.4}&\underline{89.8}&{91.9}&\underline{84.1}&92.8&97.7&98.7\\
&{DeMo}~\cite{wang2024demo}& {AAAI25}&CLIP&{79.0}&82.3&88.8&{92.0}&83.6&\underline{93.1}&\underline{97.5}&\underline{98.7}\\
&{IDEA}~\cite{wang2025IDEA}& {CVPR25}&CLIP&\underline{80.2}&{82.1}&{90.0}&\textbf{93.3}&-&-&-&-\\
&{\textbf{MGRNet$^\dagger$}}& \textbf{Ours}&CLIP&
\textbf{80.5}&\textbf{85.0}&\textbf{90.0}&\underline{92.6}&\textbf{84.6}&\textbf{93.6}&\textbf{97.7}&\textbf{98.8}\\
\bottomrule
\end{tabular}
}
\end{table*}

\subsection{Dataset and Evaluation Metrics}
We utilize four commonly multi-modal ReID datasets to evaluate our MGRNet, including two-person ReID datasets (RGBNT201~\cite{Zheng2021RobustMP},  Market1501-MM~\cite{Wang2022InteractEA}) and two-vehicle ReID datasets (RGBNT100~\cite{Li2020MultiSpectralVR} and MSVR310~\cite{ZHENG2023101901}).
RGBNT201~\cite{Zheng2021RobustMP} has 201 identities with four different viewpoints, with varying lighting and background complexity challenges.
Market1501-MM~\cite{Wang2022InteractEA} is a virtual multi-modality dataset generated from a single-modality dataset~\cite{7410490} via the cycle-GAN method~\cite{zhu2017gan} for person ReID, yielding 1501 identities with reducing 60\% of the brightness.
RGBNT100~\cite{Li2020MultiSpectralVR} is a large-scale dataset, comprising 17250 images triples of 100 vehicles across RGB, NIR and TIR modalities. In contrast, MSVR310~\cite{ZHENG2023101901} is a smaller-scale dataset that includes more complex scenarios.
For evaluation metrics, we adopt the previous evaluation strategy, utilizing mean Average Precision (mAP) and Cumulative Matching Characteristics (CMC) at Rank-K (K = 1, 5, 10) for all used datasets. Higher values of these metrics imply better model performance.

\begin{table*}[!ht]
\caption{ Comparison with  state-of-the-art methods in missing modality on the RGBNT201 dataset. M($\cdot$) denotes 
the $\cdot$ modal absence.}
\label{mytable2}
\centering
\resizebox{2.05\columnwidth}{!}{
\begin{tabular}{c|c|cc|cc|cc|cc|cc|cc}
\toprule
\multirow{2}{*}{\textbf{Methods}}&\multirow{2}{*}{\textbf{Structure}}
&\multicolumn{2}{c|}{\textbf{{M(R)}}} 
&\multicolumn{2}{c|}{\textbf{{M(N)}}}
&\multicolumn{2}{c|}{\textbf{{M(T)}}}
&\multicolumn{2}{c|}{\textbf{{M(RN)}}}
&\multicolumn{2}{c|}{\textbf{{M(RT)}}}
&\multicolumn{2}{c}{\textbf{{M(NT)}}}\\
\cmidrule{3-14}
&&\textbf{mAP} & \textbf{R-1}&\textbf{mAP} & \textbf{R-1}&\textbf{mAP} & \textbf{R-1}&\textbf{mAP} & \textbf{R-1}&\textbf{mAP} & \textbf{R-1}&\textbf{mAP} & \textbf{R-1}\\
\midrule
{HACNN~\cite{Li_2018_CVPR}}&CNN&12.5&11.1&20.5& 19.4& 16.7 &13.3& 9.2 &6.2 &6.3& 2.2 &14.8& 12.0 \\
{MUDeep~\cite{Qian2017MultiscaleDL}}&CNN&19.2&16.4&20.0& 17.2& 18.4& 14.2& 13.7& 11.8& 11.5& 6.5 &12.7 &8.5\\
{OSNet~\cite{9011001}}&CNN&19.8&17.3&21.0 &19.0& 18.7& 14.6& 12.3& 10.9&9.4& 5.4& 13.0& 10.2\\
{MFLN~\cite{8578323}}&CNN&20.2&18.9&21.1& 19.7& 17.6 &11.1& 13.2 &12.1 &8.3 &3.5& 13.1& 9.1\\
{CAL~\cite{Rao2021CounterfactualAL}}&CNN&21.4 &22.1 &24.2& 23.6& 18.0 &12.4& 18.6& 20.1& 10.0& 5.9& 17.2& 13.2\\
{PCB~\cite{Sun2017BeyondPM}}&CNN&23.6& 24.2 & 24.4& 25.1& 19.9 &14.7 &20.6 &23.6& 11.0 &6.8& 18.6& 14.4\\
{PFNet~\cite{Zheng2021RobustMP}}&CNN&-&-&31.9&29.8&25.5&25.8&-&-&-&-&26.4& 23.4\\
{DENet~\cite{Zheng2023DynamicEN}}&CNN&-&-&35.4&36.8&33.0&35.4&-&-&-&-&32.4 &29.2\\
{TOP-ReID~\cite{Wang2023TOPReIDMO}}&ViT&{54.4}&{57.5}& {64.3}&{67.6}&{51.9}&{54.5}& {35.3}& {35.4}&{26.2}&{26.0}&{34.1}& {31.7}\\
{DeMo~\cite{wang2024demo}}&CLIP&\underline{63.3}&\underline{65.3}& \underline{72.6}&\underline{75.7}&\underline{56.2}&{54.1}& \textbf{45.6}&\textbf{46.5}&{26.3}&{24.9}&\underline{40.3}&\underline{38.5}\\
\midrule
{\textbf{{MGRNet$^\ast$}}}&ViT&{54.6}&{54.1}& {68.4}  &{70.7}&{55.8}&\underline{57.1}&{36.4}&{36.4}&\underline{26.8}&\underline{27.4}&{36.2}&{32.4}\\
{\textbf{MGRNet$^\dagger$}}&CLIP&\textbf{66.3}&\textbf{68.8}& \textbf{75.2} &\textbf{78.1}&\textbf{58.6}&\textbf{59.0}&\underline{44.1}&\underline{43.7}&\textbf{30.7 }&\textbf{29.8}&\textbf{42.6}&\textbf{42.0}\\
\bottomrule
\end{tabular}
}
\end{table*}
\begin{table}[!ht]
\caption{Comparisons with state-of-the-art methods on vehicle ReID datasets.}
\label{mytable0}
\centering
\resizebox{1.0\columnwidth}{!}{
\begin{tabular}{c|c|c|cc|cc}
\toprule
&\multirow{2}{*}{\textbf{{Methods}}}&\multirow{2}{*}{\textbf{{Structure}}}
&\multicolumn{2}{c|}{\textbf{{RGBNT100}}} 
&\multicolumn{2}{c}{\textbf{{MSVR310}}}\\
\cmidrule{4-5}
\cmidrule{6-7}
&&&\textbf{mAP} & \textbf{R-1}&\textbf{mAP} & \textbf{R-1}\\
\midrule
\multirow{10}{*}{\rotatebox{90}{\textbf{Single}}}
&{PCB}~\cite{Sun2017BeyondPM}&CNN&57.2& 83.5 &23.2 &42.9 \\
&{MGN}~\cite{Wang2018LearningDF}&CNN&58.1 &83.1 &26.2& 44.3 \\
&{DMML}~\cite{9009474}&CNN&58.5& 82.0 &19.1 &31.1 \\
&{HRCN}~\cite{9711371}&CNN&67.1& 91.8& 23.4& 44.2 \\
&{AGW}~\cite{9336268}&CNN&73.1 &92.7& 28.9& 46.9\\
&{OSNet}~\cite{9011001}&CNN&75.0& 95.6& 28.7& 44.8\\
&{BoT}~\cite{9025455}&CNN & 78.0& 95.1 &23.5& 38.4 \\
&{TransReID}~\cite{He_2021_ICCV}&ViT&75.6 &92.9& 18.4& 29.6 \\
\midrule
\multirow{16}{*}{\rotatebox{90}{\textbf{Multi}}}
&{PFNet}~\cite{Zheng2021RobustMP}&CNN&68.1& 94.1& 23.5& 37.4\\
&{IEEE}~\cite{Wang2022InteractEA}&CNN&61.3 &87.8 &21.0 &41.0 \\
&{GAFNet}~\cite{Guo2022GenerativeAA}&CNN&74.4 &93.4& -&-\\
&{HAMNet}~\cite{Li2020MultiSpectralVR}&CNN&74.5& 93.3 &27.1& 42.3\\
&{CCNet}~\cite{ZHENG2023101901}&CNN&77.2& 96.3 &36.4& \textbf{55.2}\\
&{GraFT}~\cite{Yin2023GraFTGF}&ViT&76.6&94.3& -& -\\
&{HTT}~\cite{Wang2024HeterogeneousTT}&ViT&75.7& 92.6& -& -\\
&{TOP-ReID}~\cite{Wang2023TOPReIDMO}&ViT& 81.2& 96.4& 35.9& 44.6 \\
&{FACENet}~\cite{ZHENG2025102800}&ViT& 81.5& \underline{96.9}& 36.2 &\underline{54.1} \\
&{EDITOR}~\cite{10654953}&ViT&82.1&96.4&39.0&49.3 \\
&{RSCNet}~\cite{10772090}&ViT&82.3&96.6&\underline{39.5}&{49.6} \\
&{DeMo}~\cite{wang2024demo}&ViT&\underline{82.4} & 96.0 &{39.1} & {48.6}  \\
&{\textbf{MGRNet$^\ast$}}&ViT&\textbf{83.0}	&\textbf{97.1}&\textbf{39.9}&{49.3}\\
\cmidrule{2-7}
&{MambaPro}~\cite{Wang2024MambaPro}&CLIP&83.9 & 94.7 &  47.0 & 56.5 \\
&{DeMo}~\cite{wang2024demo}&CLIP&86.2 & \textbf{97.6}  &\underline{49.2} & {59.8}  \\
&{IDEA}~\cite{wang2025IDEA}&CLIP&87.2 & 96.5&47.0&\underline{62.4} \\
&{\textbf{MGRNet$^\dagger$}}&CLIP&\textbf{87.8}	&\underline{97.3}&\textbf{53.2}&\textbf{67.2}\\
\bottomrule
\end{tabular}
}
\end{table}
\subsection{Implementation details}
The proposed model is implemented via Pytorch with one RTX 4090 GPU. 
For all used datasets, we resize images into 256 × 128 × 3 pixels for person datasets and 128 × 256 × 3 pixels for vehicle datasets. Furthermore, we adopt previous random erasure, random flipping, and padding in the training~\cite{He_2021_ICCV}. Meanwhile, the batch size in the training phase is set to 64 including 4 different identities. the number of epochs is set to 80/40 and the SGD/Adam~\cite{DBLP:journals/corr/KingmaB14} optimizer is used for the ViT/CLIP-based MGRNet, with a momentum coefficient of 0.9 and a weight decay of 0.0001. The initial learning rate is set to 0.0066/0.00035, following a warmup strategy combined with cosine decay.

\subsection{Comparisons with State-of-the-Art Methods}
We conduct comprehensive comparisons with state-of-the-art methods, including  CNN-, ViT- and CLIP-based methods.
In general, CNN-based methods often tend to be lower as shown in Table~\ref{mytable1} and~\ref{mytable0}, showcasing the effectiveness of ViT and CLIP for multi-modal fusion strategy. 
For a fair and
comprehensive comparison, we evaluate our proposed method using different vision encoders (e.g., ViT and CLIP) on four public benchmarks for multi-modal object ReID.
\textbf{Comparison on RGBNT201 and Market1501-MM. }
As shown in Table~\ref{mytable1}, One observation is that we achieve excellent performance compared to other methods for both ViT-based and CLIP-based MGRNet on two-person datasets.
Notably, 
DeMo~\cite{wang2024demo}  achieves promising performance by considering the global information of each modality with the local information of all modalities.
Meanwhile, IDEA~\cite{wang2025IDEA} fusions semantic information to generated sampling local information, improving the identification ability.
However, these methods treat all local patches equally and fail to distinguish patches of different quality.
Compared with them, 
our approach can effectively pay attention to the quality differences of patches,  extracting more discriminating features of persons.
Eventually, for the RGBNT201 dataset~\cite{Zheng2021RobustMP}, the ViT-based MGRNet
notable performance  with scores of 78.4\%, 82.5\%, 90.9\% and 95.2\% over the sub-optimal ViT-based method by 4.7\%, 2.0\%, 2.6\%, 3.7\% for mAP, Rank-1, Rank-5 and Rank-10, respectively. Also, although IDEA~\cite{wang2025IDEA} introduces supplementary semantic information to improve identification ability, our CLIP-based MGRNet consistently achieves higher results in mAP and Rank-1, demonstrating the robustness of its multi-modal representation learning.
For larger-scale generated Market1501-MM of multi-modal data~\cite{Wang2022InteractEA}, our method outperforms the sub-optimal results in mAP and Rank-1, which are higher than 2.4\%/0.5\%, and 1.6\%/0.5\% 
 for the ViT/CLIP-based MGRNet, respectively. 
The performance improvement is relatively modest compared with RGBNT201~\cite{Zheng2021RobustMP}, likely since 
it is a virtual multi-modal person ReID dataset generated by a GAN network~\cite{zhu2017gan}. As a result, it presents additional challenges during training.

\textbf{Comparison on RGBNT100 and MSVR310. } 
To further validate our proposed method, related vehicle experiments are performed.
Table~\ref{mytable0} shows that MGRNet obtains
relatively superior performance for different vision encoders. Our results find that the ViT-based MGRNet achieves better performance than the latest ViT-based DeMo. Moreover, the CLIP-based variant consistently outperforms the semantic-guide CLIP-based IDEA, with gains of 0.6\%/0.8\% on RGBNT100~\cite{Li2020MultiSpectralVR} and 6.2\%/4.8\% on MSVR310~\cite{ZHENG2023101901} in mAP/Rank-1, respectively.
These results indicate the effectiveness of our approach by considering the quality difference of local features and reducing the influence of low-quality local features.
They also confirm our proposed model's ability to effectively extract and integrate discriminative cues from the ReID task in complex and heterogeneous data environments.

\subsection{Evaluation on Missing modality}
Table~\ref{mytable2} evaluates the proposed GRMM strategy, which presents the experimental results in simulating missing modality scenarios. Our model consistently delivers strong performance.
One is observed that our results outperform methods based on feature reconstruction TOPReID~\cite{He_2021_ICCV} and the latest DeMo~\cite{wang2024demo}.
The average performance is higher by 2.0\%/2.2\% and 0.9\%/2.9\% in mAP and Rank-1 for ViT/CLIP-based methods. Meanwhile, we find that even in the absence of RGB/NIR modality, our results are still better than the single-modal method as well as some multi-modal methods, especially EDITOR~\cite{10654953}.
These indicate that graph reasoning can effectively reconstruct missing modal information by considering the essential structural relationships, yielding reliable multi-modality data in object ReID.

\begin{table*}[!ht]
\caption{Comparison results for different modules on the RGBNT201 dataset.}
\label{ablation}
\centering
\resizebox{2.04\columnwidth}{!}{
\begin{tabular}{c c c c c | c c c c |c c c c}
\toprule
\multicolumn{5}{c|}{\textbf{Modules}}
&\multicolumn{4}{c|}{\textbf{ViT}} &\multicolumn{4}{c}{\textbf{CLIP}}\\
\midrule
& \textbf{MGL + LGR} & \textbf{SGNS$^{1st}$} &\textbf{SGNS$^{2nd}$} & \textbf{GRMM} & \textbf{mAP} & \textbf{R-1} & \textbf{R-5} & \textbf{R-10} & \textbf{mAP} & \textbf{R-1} & \textbf{R-5} & \textbf{R-10}\\
\midrule
(a)&$\times$ & $\times$ & $\times$ & $\times$ & 74.2 & 78.1 & 89.0 & 92.7& 73.1& 76.4 &84.7 &89.2 \\
(b)&\checkmark & $\times$ & $\times$ & $\times$ & 75.0 & 78.5 & 88.8 & 93.1& 74.6 & 79.1 & 86.7 & 90.9 \\
(c)&\checkmark & \checkmark & $\times$ & $\times$ & 75.7 &79.5& 90.8 &94.0& 75.8& 80.3& 87.9 &91.0  \\
(d)&\checkmark & \checkmark& \checkmark & $\times$ & 76.9 & 80.9 & 90.9 & 93.8 & 78.3 & 82.1 & 87.9 & 91.4 \\
(e)&\checkmark & $\times$ & $\times$ & \checkmark & 77.2 & 80.5 & 88.9 & 93.5 & 75.2 &79.3& 89.2 &91.6 \\
(f)&\checkmark & \checkmark & $\times$ & \checkmark & 78.0 & 81.0 & 88.5 & 92.6 & 79.9 &83.3 &89.5 &92.1 \\
\midrule
(g)&\checkmark & \checkmark & \checkmark & \checkmark & \textbf{78.4} & \textbf{82.5} & \textbf{90.9} & \textbf{95.2} & \textbf{80.5} & \textbf{85.0} & \textbf{90.0} & \textbf{92.6} \\
\bottomrule
\end{tabular}
}
\end{table*}

\subsection{Ablation Study}
To assess the effectiveness of the modules in our proposed MGRNet, we establish a baseline model as Table~\ref{ablation} (a) that consists of the multi-branch vision encoders ViT/CLIP, Global-aware Multi-Head Attention, and the relevant optimization losses. We incrementally incorporate our proposed components into the baseline. 
\begin{figure}
\centering
\includegraphics[width=1\columnwidth]{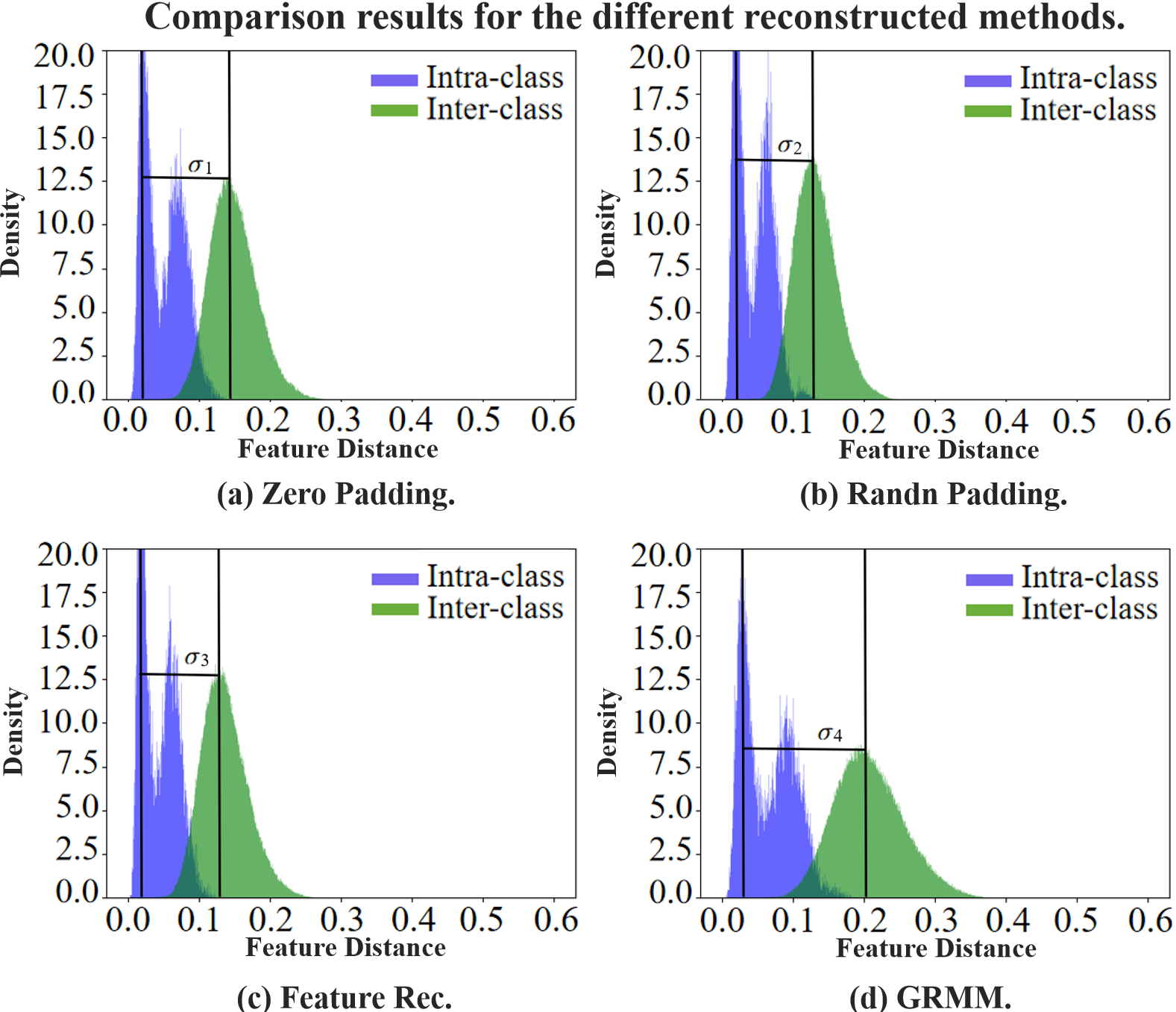} 
\caption{
The intra-class and inter-class distances of cross-modality features of different methods.}
\label{Re}
\end{figure}
\begin{figure}
\centering
\includegraphics[width=1\columnwidth]{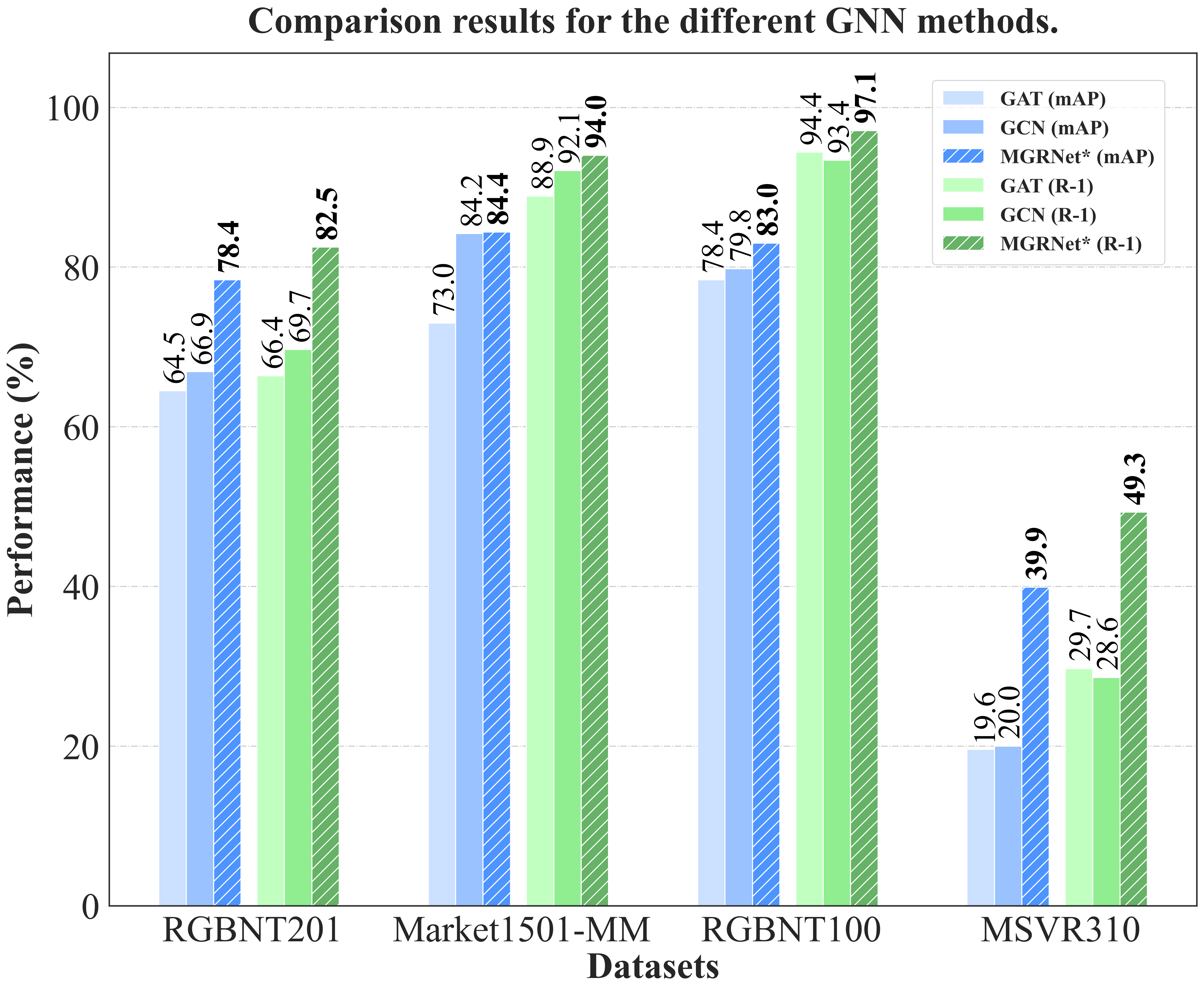} 
\caption{
Comparison results for GNN methods on the common dataset.}
\label{GNN}
\end{figure}
 
As shown in Table~\ref{ablation} (b), the integration of Modality-aware Graphs Learning (MGL) and Local-aware Graph Reasoning (LGR)  enables the model to focus on critical local details for multi-modal data, obtaining a performance improvement in mAP and Rank-1, respectively.
Furthermore, in Table~\ref{ablation} (d), the incorporation of the Selective Graph Nodes Swap (SGNS) significantly enhances performance, yielding a 2.7\%/5.2\% and 2.8\%/5.7\% improvement in mAP and Rank-1 compared to the ViT/CLIP-based baseline. 
This observation suggests that SGNS effectively mitigates the impact of multi-modal local noise. 
Lastly, as shown in Table~\ref{ablation} (e), our proposed Graph Reasoning on Missing Modality (GRMM) leverages both feature and structural relationships to restore modality-specific representations effectively. 
This process not only reduces modality discrepancies but also further boosts mAP and Rank-1 performance compared with Table~\ref{ablation} (b). 
Additionally, to evaluate the effectiveness of each step in the SGNS operation in Fig. \ref{part}, we further analyze the two selection operations. 
As can be seen from the experimental results in Table \ref{ablation} (c) and (f), our proposed SGNS operation is effective and can gradually alleviate the impact and noise caused by low-quality local features, thus further improving the model performance for multi-modal ReID tasks.

Summarily, integrating these components into the baseline model yields a notable improvement in experimental performance. The performance gains observed across various evaluation metrics validate the effectiveness of our proposed MGRNet. These results further emphasize the importance of structured graph-based fusion techniques in the field. 

\begin{table*}[!ht]
\caption{Comparison results for different methods of Reconstruction on the RGBNT201 dataset.}
\label{mytable5}
\centering
\scalebox{1.2}{
\begin{tabular}{c|cccc|cccc|cccc}
\toprule
{\multirow{2}{*}{\textbf{Modules}}} 
&\multicolumn{4}{c|}{\textbf{ALL}}
&\multicolumn{4}{c|}{\textbf{M(N)}} 
&\multicolumn{4}{c}{\textbf{M(RT)}}
\\
\cmidrule{2-5}
\cmidrule{6-9}
\cmidrule{10-13}
&\textbf{mAP} & \textbf{R-1}&\textbf{R-5} & \textbf{R-10}&\textbf{mAP} & \textbf{R-1}&\textbf{R-5} & \textbf{R-10}&\textbf{mAP} & \textbf{R-1}&\textbf{R-5} & \textbf{R-10}\\
\midrule
{Zero Padding.~\cite{9156763}}&74.4&76.7& 87.0	&92.1&67.0&67.7&81.1&85.9&23.3&21.8&35.6&46.7 \\
{Random Padding.~\cite{9156763}}&76.3&79.5& 88.6	&92.9&66.8&69.4&81.0	&88.3&24.3&22.4& 39.2	&51.9\\
{Feature Rec.~\cite{Wang2023TOPReIDMO}}&76.4&78.7& 89.4  &94.5&66.1&69.0& 80.9&	88.6&25.7&27.3& 42.8	&53.9\\
\midrule
\textbf{MGRNet$^\ast$}&\textbf{78.4}&\textbf{82.5}& \textbf{90.9}  &\textbf{95.2}&\textbf{68.4}&\textbf{70.7}& \textbf{81.8}  &\textbf{88.8}&\textbf{26.8}&\textbf{27.4}& \textbf{43.5}  &\textbf{55.1}\\
\bottomrule
\end{tabular}
}
\end{table*}

\begin{figure}[t]
\centering
\includegraphics[width=1.02\columnwidth]{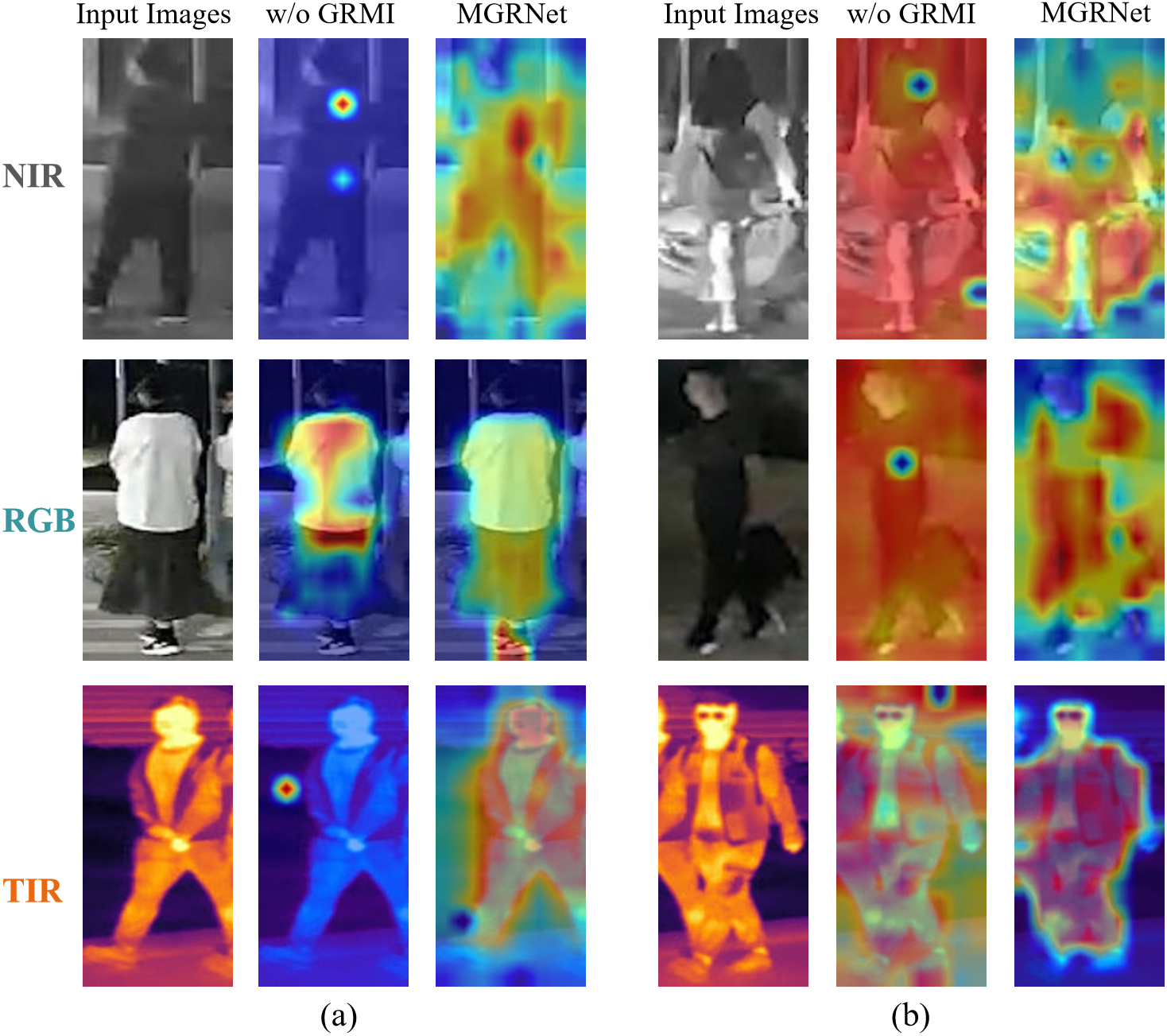}
\caption{
Visualization results using Gradient-weighted Class Activation Mapping (Grad-CAM) of models without and with Graph Reasoning on Modal Interaction (GRMI).
}
\label{Visualization}
\end{figure}

\begin{figure}[t]
\centering
\includegraphics[width=1\columnwidth]{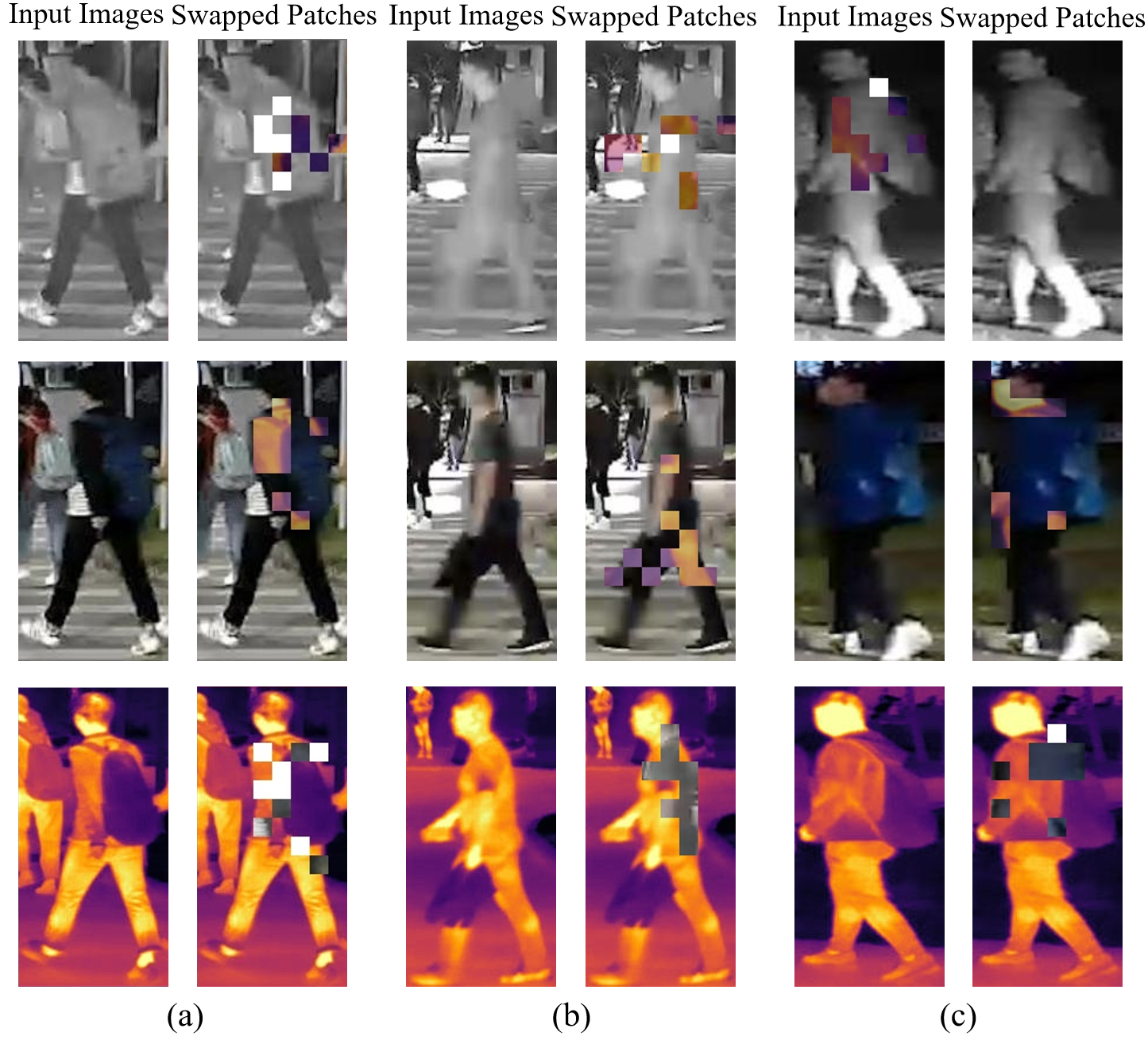} 
\caption{
Visualization results of swapped patches of Graph Reasoning on Modal Interaction (GRMI)  on the RGBNT201 dataset. 
}
\label{swap}
\end{figure}

\subsection{Evaluation on GRMM}
We conduct several experiments using common reconstruction techniques to validate the proposed GRMM strategy. By modifying different reconstruction methods, we can assess the contribution of our strategy to the overall performance for complete modalities and missing different modalities in Table~\ref{mytable5}. The results demonstrate that the proposed GRMM strategy consistently outperforms other configurations, confirming its critical role in enhancing the model’s performance. Meanwhile, we visualize the inter-class and intra-class distances for different reconstruction methods as depicted in Fig.~\ref{Re}, where ${\sigma}_4 > {\sigma}_1, {\sigma}_2, {\sigma}_3$.   This shows that the intra-class distance of GRMM is significantly reduced compared with other reconstruction methods.

\subsection{Evaluation on GNN}
We conduct the comparative experiments with standard GNN approaches~\cite{veličković2018graph,Kipf2017}, including Graph Attention Network (GAT)~\cite{veličković2018graph}, Graph Convolutional Network (GCN)~\cite{Kipf2017}. These results are summarized in Fig.~\ref{GNN}. The approaches first employ non-shared multi-branch ViT to extract multi-modal features, and then construct a graph for all patches of each modality via Euclidean Distance~\cite{10.1145/3459637.3482261}. Finally, both label smoothing cross-entropy and triplet loss are combined to optimize the entire network.
The results demonstrate that MGRNet consistently outperforms these traditional methods. This enhanced performance can be attributed to MGRNet's ability to alleviate the impact of low-quality local features, enhancing the discriminative information. 

\begin{table}[!ht]
\caption{Comparison results for the number of swap-nodes on the RGBNT201 dataset.}
\label{mytable7}
\centering
\scalebox{1}{
\begin{tabular}{c|cccc|cccc}
\toprule
&\multicolumn{4}{c|}{\textbf{ViT}} &\multicolumn{4}{c}{\textbf{CLIP}}\\
\midrule
{$k$}&\textbf{mAP} & \textbf{R-1}&\textbf{R-5} & \textbf{R-10}&\textbf{mAP} & \textbf{R-1}&\textbf{R-5} & \textbf{R-10}\\
\midrule
0&77.2&80.5& 88.9&93.5 &75.2& 79.3 &89.2& 91.6 \\
10&78.3&82.2& 90.9&94.7&79.9&82.8& 90.2&92.6 \\
20&\textbf{78.4}&\textbf{82.5}& \textbf{90.9}&\textbf{95.2}&\textbf{80.5}&\textbf{85.0 }& 90.0&92.6  \\
40&78.0&80.3& 89.5&95.2&78.6 &82.5 &\textbf{90.7} &\textbf{92.9}\\
60&77.0&79.4&90.1&94.9&76.8&81.0&87.3&89.7\\
80&61.6&60.3& 76.2&86.0&76.6&80.0& 87.8&91.4\\
\bottomrule
\end{tabular}
}
\end{table}

\begin{figure*}[t]
\centering
\includegraphics[width=2.08\columnwidth]{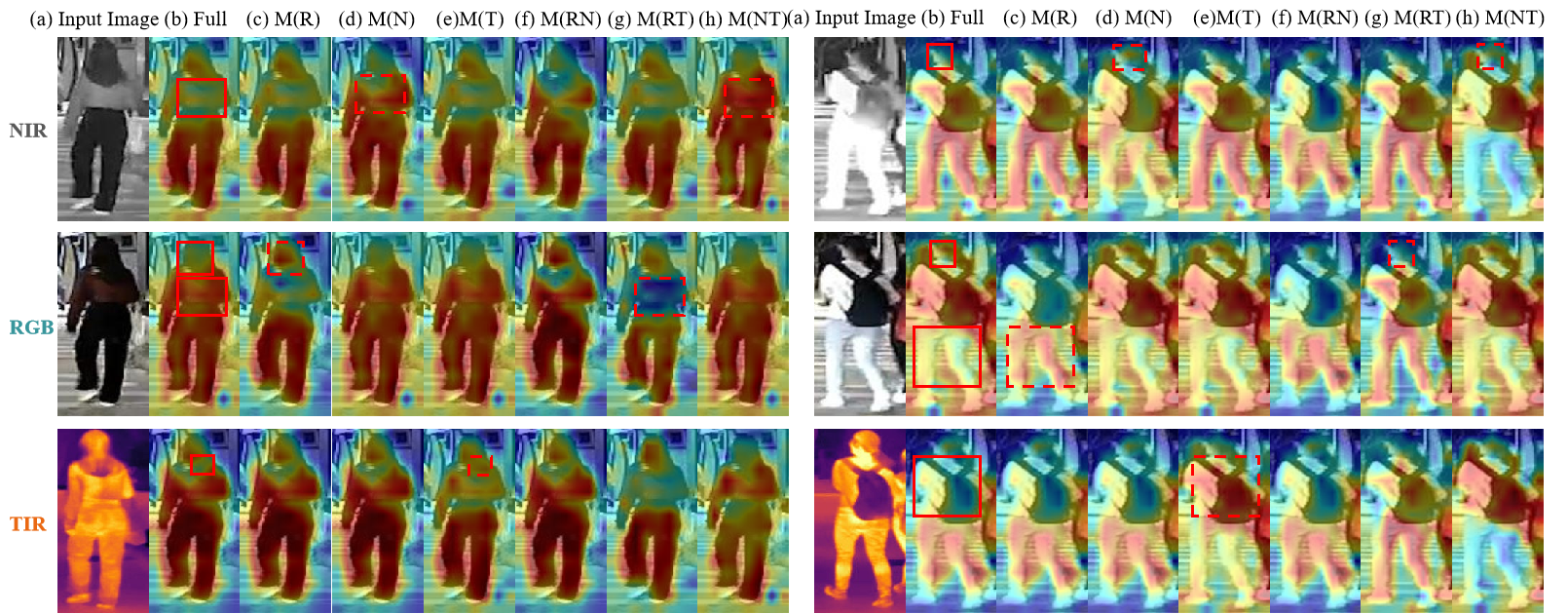}
\caption{Feature map for complete and missing modality on the RGBNT201 dataset.}
\label{person_miss}
\end{figure*}

\subsection{Hyperparameter Analysis} 
We further analyze the influence of the swap-node number (hyperparameter 
$k$) on the performance of our model.
The results are displayed in Table~\ref{mytable7}, where the first row represents the case where the SGNS method is not applied in different vision encoders.
From the results, we observe a trend: as the  $k$-value increases, the model’s performance initially improves before reaching a peak and then declines. This behavior highlights that an optimal number of swapped nodes enhances the model's ability to capture rich local features and reduce the impact of low-quality local features. These results prove that our SGNS operation is more effective than using a direct graph-based approach.

\subsection{Visualization}
\textbf{Interaction Feature.}
We present visualization with and without the GRMI strategy, utilizing Grad-CAM~\cite{8237336}, which demonstrates the ability of our MGRNet to effectively capture the relevant regions of the input images. Compared to the proposed MGRNet without the GRMI strategy, our method can capture more critical areas while significantly reducing the impact of irrelevant or noisy regions. 
The results in Fig.~\ref{Visualization} (a) show that our method can capture key information more comprehensively, while Fig.~\ref{Visualization} (b) shows that this method has a better effect in background noise suppression, thus improving the quality of overall feature representation.

Additionally, to further evaluate the GRMI strategy, we visualize the exchanged patches via the SGNS operation, as shown in Fig. \ref{swap}. By exchanging local features, problems related to low-quality features, such as the face and bag, are solved, enhancing the representation of discriminative information and improving the effectiveness of modal representation.

\begin{figure}[t]
\centering
\includegraphics[width=1\columnwidth]{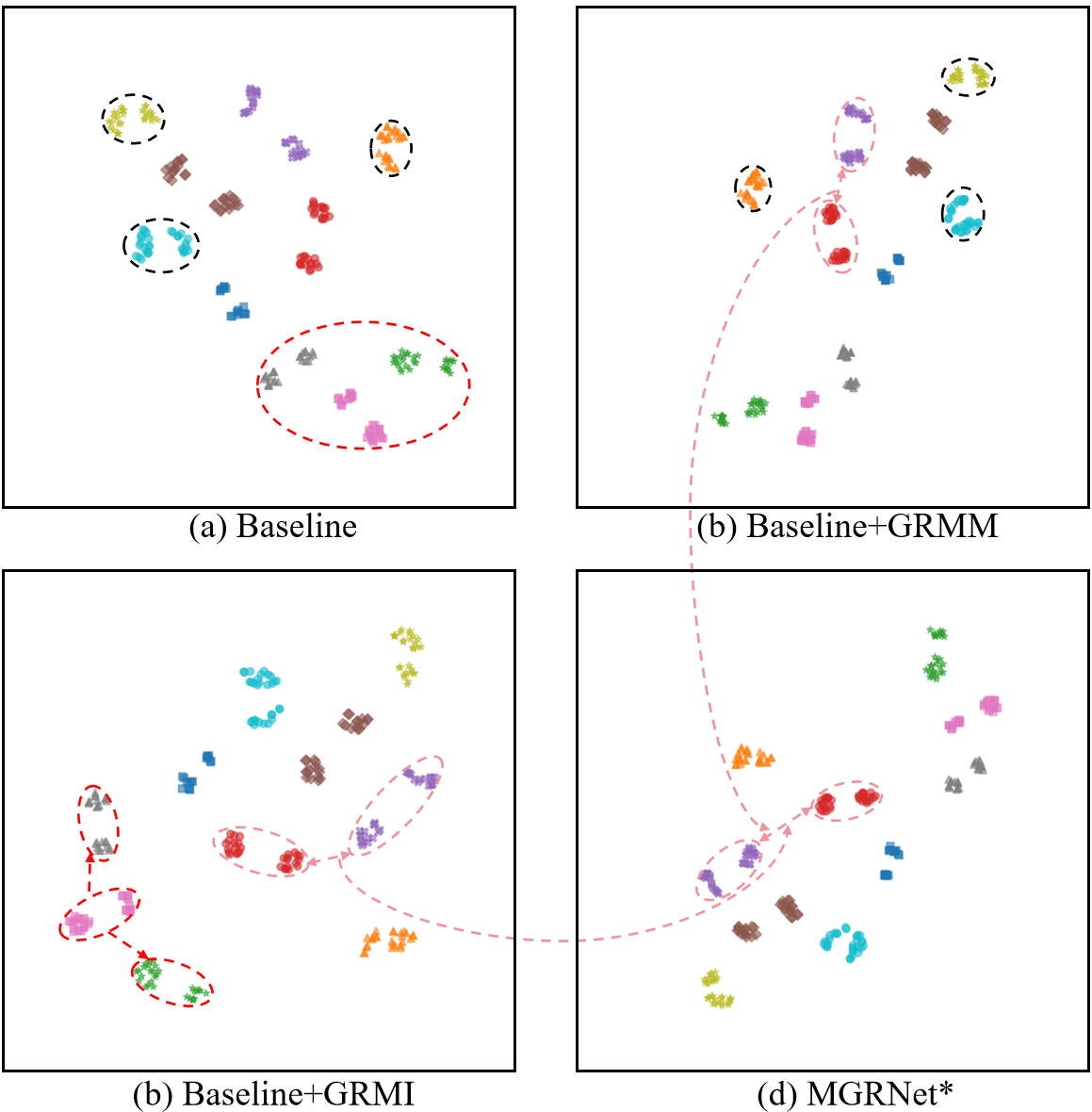} 
\caption{Feature distribution on different strategies by using t-SNE. The different colors represent different identities.}
\label{T-sne}
\end{figure}

\textbf{Reconstruction Feature.}
To more intuitively assess the GRMM strategy, we utilize feature maps to visualize different modal images. As shown in Fig.~\ref{person_miss}, we can observe that our model can recover superior feature representations in the case of missing modalities.
It is known that RGB contains more color and detail information, covering the brightness and edge features of NIR, while NIR is more prominent in low light structure and contrast and is a complement to RGB~\cite{9301787}.
One is found that the associated region in the NIR modality when generating the RGB modality is less accurate than the real one as Fig.~\ref{person_miss} (g), while the associated region in the RGB modality when generating the NIR modality is more accurate than real one as Fig.~\ref{person_miss} (h). In the future, we will consider the specific information and complementary information between modalities to further optimize our work.

\textbf{Feature Distribution.} 
We visualize the obtained representations by the t-SNE tool 
\cite{Kobak2021InitializationIC}. Fig. \ref{T-sne} (a) is the visualized result by leveraging the vit-based baseline. Fig. \ref{T-sne} (b) and (c) add our proposed GRMM and GRMI strategies into this baseline, respectively. One can observe that GRMM can reduce the distance between intra-class by minimizing the disparity between modalities, and GRMI can increase the distance between inter-class by reducing the impact of low-quality patches.   Moreover, in Fig. \ref{T-sne} (d), it can be found that our proposed MGRNet exhibits larger inter-class distances and smaller intra-class distances than the baseline, indicating that our model is better at capturing differences between objects of different modalities.
\section{Conclusion}
In this paper, we propose a novel approach named Modality-aware Graph Reasoning Network (MGRNet), effectively boosting information interactions and recovering missing modalities for multi-modal object ReID. First, we introduce the construction of modality-aware graphs to adaptively model the relationships among local features, learning important local details. 
Second, the selective graph nodes swap operation is introduced to alleviate the impact of low-quality features from the modalities while effectively capturing crucial local details, promoting multi-modal fusion. 
Finally, we seed the swapped modality-aware graphs into the local-aware graph reasoning module to achieve message propagation, thus yielding a reliable feature representation of multi-modality data in object ReID.
Additionally, we propose that the MGRNet is capable of recovering missing modalities based on their structural relationships, effectively reducing and minimizing multi-modality disparity.
Overall, the proposed MGRNet achieves state-of-the-art performance on multi-modal ReID datasets. 
MGRNet serves as a preliminary framework for multi-modal fusion and recovery in graph reasoning and demonstrates promising results. However, multi-modal fusion remains a challenging task. In the future, our works will focus on developing more efficient and fine-grained fusion strategies of graph reasoning.

\bibliographystyle{IEEEtran}
\bibliography{ieee_mgrnet}{}

\newpage

\vfill

\end{document}